\documentclass{article}

\usepackage{PRIMEarxiv}

\usepackage[utf8]{inputenc}
\usepackage[T1]{fontenc}
\usepackage{hyperref}
\usepackage{url}
\usepackage{booktabs}
\usepackage{amsfonts}
\usepackage{amssymb}
\usepackage{amsmath}
\usepackage{nicefrac}
\usepackage{microtype}
\usepackage{graphicx}
\usepackage{multirow}
\usepackage{array}
\usepackage{xcolor}
\usepackage{verbatim}
\usepackage{enumitem}
\usepackage{tikz}
\usetikzlibrary{positioning,arrows.meta}
\usepackage[capitalize,noabbrev]{cleveref}

\newcommand{\best}[1]{\underline{#1}}

\title{
    \includegraphics[width=\textwidth]{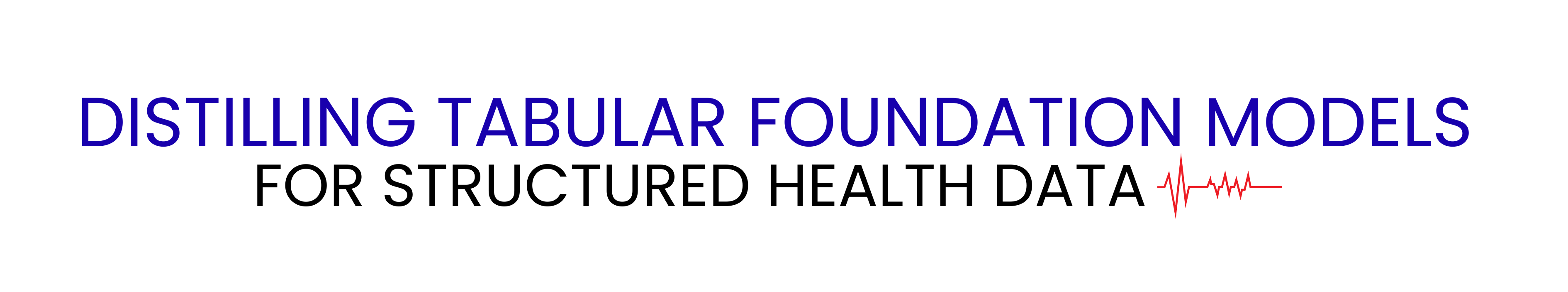}
}

\author{
  Aditya Tanna, 
  Nassim Bouarour, 
  Mohamed Bouadi, \\
  Vinay Kumar Sankarapu, 
  Pratinav Seth \\
  \affiliation{Lexsi Labs} \\
}

\setabstract{%
Tabular foundation models (TFMs) achieve strong performance on health datasets, but their inference cost and infrastructure requirements limit practical use. We study whether their predictive behavior can be transferred to lightweight tabular models through knowledge distillation. Since in-context TFMs condition on the training set at inference time, naive distillation can introduce context leakage; we address this with stratified out-of-fold teacher labeling. Across $19$ healthcare datasets, $6$ TFM teachers, $4$ student families, and several multi-teacher ensembles, we find that distilled students retain at least $90\%$ of teacher AUC, outperforming teachers in some cases, while running at least $26\times$ faster on CPU and preserving calibration and fairness critical for health applications. Moreover, multi-teacher averaging does not consistently improve over the best single teacher. Leakage-aware distillation is thus a viable route for bringing TFM-quality predictions into inference-constrained health settings.
}

\setkeywords{knowledge distillation, tabular foundation models, clinical AI, structured health data, model compression, calibration, fairness}

\runningtitle{Distilling Tabular Foundation Models for Structured Health Data}

\begin{document}
\maketitle

\section{Introduction}
\label{sec:introduction}

Structured data is central to health machine learning~\cite{rajpurkar2022ai}. Tasks such as risk stratification, disease screening, readmission prediction, and mortality estimation are often defined over EHR, laboratory values, demographics, and other tabular variables~\cite{johnson2016mimic}. These datasets are frequently small, heterogeneous, and sparse, making data-efficient prediction essential. Healthcare deployment also requires more than high accuracy. Hospital systems are often CPU-based, air-gapped, and constrained by data-residency policies, limiting the use of cloud-GPU inference. Models must produce calibrated probabilities that support clinical decisions~\cite{guo2017calibration} and avoid amplifying demographic disparities that can affect access to care~\cite{obermeyer2019dissecting,hardt2016equality}.

Tabular foundation models (TFMs) have recently emerged as a promising approach for such clinical datasets. Models such as TabPFNv2.5~\cite{grinsztajn2025tabpfn25}, TabICLv2~\cite{qu2025tabiclv2}, TabDPT~\cite{ma2025tabdpt}, LimiX~\cite{zhang2025limix}, and Orion-MSP~\cite{bouadi2025orionmsp} often match or outperform tuned gradient-boosted trees on datasets with fewer than ten thousand rows, without per-task tuning.

However, the TFM inference paradigm conflicts with the practical constraints of clinical deployment. Predictions require conditioning on a task-specific context set and running large neural models on dedicated GPU hardware. Many hospital systems are CPU-based, air-gapped, or constrained by data-residency rules, and these requirements rule out direct deployment. Even where GPU inference is available, latency on the order of hundreds of milliseconds limits the use of TFMs in high-throughput batch scoring (whole-hospital risk surveillance, registry sweeps, retrospective audits) and in real-time alert pipelines that must integrate with existing monitoring and auditing infrastructure.

We therefore ask whether TFMs' predictive behavior, including accuracy, calibration, and fairness, can be transferred to standard ML models that are easier to deploy. We study this question through knowledge distillation~\cite{hinton2015distilling} from TFMs into lightweight students. Since TFM teachers condition on labeled training examples at inference time, teacher outputs for examples included in the context can produce biased soft targets~\cite{mansurov2024datalaundering}. We mitigate this using stratified out-of-fold teacher labeling. We run extensive experiments across $19$ health datasets ($16$ binary, $3$ multiclass), using $6$ TFMs as teachers, $4$ standard models---LightGBM~\cite{ke2017lightgbm}, CatBoost~\cite{prokhorenkova2018catboost}, XGBoost~\cite{chen2016xgboost}, and an MLP---as students, and $4$ multi-teacher ensembles. We find that: (i) distilled students retain more than $90$\% of teacher AUC, outperforming teachers on some datasets; (ii) MLP students are $2\times$ faster than LightGBM but are miscalibrated and amplify fairness gaps; (iii) Multi-teacher settings tend to be outperformed by the best single teacher.

\paragraph{Contributions.} We summarize our contributions as follows:
\begin{itemize}[leftmargin=*, itemsep=2pt, topsep=2pt, parsep=0pt, partopsep=0pt]
    \item We study knowledge distillation as a practical route for transferring TFM predictions into deployable healthcare tabular models.
    \item We propose a stratified out-of-fold teacher-labeling scheme to prevent \emph{teacher identity leakage} from ICL-based TFMs.
    \item We benchmark TFM distillation across 19 healthcare datasets, 6 teachers, 4 student families, and multi-teacher ensembles.
    \item We show that tree-based students provide the strongest accuracy--calibration--fairness tradeoff, while MLP distillation and naive multi-teacher averaging are less consistently beneficial.
\end{itemize}

\section{Related Work}
\label{sec:related}

\paragraph{Tabular foundation models.} TFMs pretrain on synthetic or curated real tables and then use in-context learning at inference time. TabPFN~\cite{hollmann2025tabpfn} and its v2.5/v2.6 update~\cite{grinsztajn2025tabpfn25} are prior-fitted networks trained on synthetic Bayesian-style data. TabICLv2~\cite{qu2025tabiclv2} focuses on context length and inference throughput. TabDPT~\cite{ma2025tabdpt} pretrains on a large corpus of real tables. LimiX~\cite{zhang2025limix}, Orion-Bix~\cite{bouadi2026orionbix} and Orion-MSP~\cite{bouadi2025orionmsp} extend the approach with multi-scale or sparse attention. Despite the architectural variation, the deployment profile is essentially the same: a GPU forward pass conditioned on the labeled training set, with cost that scales with context size and memory in the tens to hundreds of megabytes on accelerator hardware.

\paragraph{Knowledge distillation.} Distillation transfers a teacher's predictive distribution to a student through softened probability targets~\cite{hinton2015distilling}. Follow-up work has examined when soft labels actually help~\cite{tang2020understanding,zhou2021rethinking} and when ensemble averaging adds noise rather than signal~\cite{du2020agree}. The technique is widely used in vision, language, and speech, but the tabular setting has received much less attention. Two recent findings are directly relevant here: soft labels can carry held-out information from the teacher, including memorised training points~\cite{behrens2025softleak}, and benchmark scores can be inflated through indirect distillation chains~\cite{mansurov2024datalaundering}.

\paragraph{In-context teachers and leakage.} Because TFM teachers condition on the training set at inference, scoring those same training points returns predictions on examples the model has effectively seen. We refer to this as \emph{teacher identity leakage}. The standard remedy in stacking is $K$-fold out-of-fold prediction, and we adopt it here. To our knowledge, the only published TFM distillation pipeline is the closed-source engine packaged with TabPFNv2.5~\cite{grinsztajn2025tabpfn25}; we cannot benchmark directly against it. Our pipeline is open-source, model-agnostic across multiple TFM families, and supports multi-teacher ensembles.

\paragraph{Clinical tabular machine learning.} Healthcare prediction tasks routinely operate on tabular inputs derived from EHR data~\cite{johnson2016mimic,rajpurkar2022ai}. Two requirements distinguish this domain from generic tabular benchmarks. Probabilities need to be calibrated so they can be plugged into threshold-based decision rules~\cite{guo2017calibration,platt1999probabilistic,naeini2015calibration}. Group fairness has to be evaluated alongside accuracy, since miscalibrated or biased models can affect care decisions~\cite{hardt2016equality,obermeyer2019dissecting}. Recent tabular fine-tuning work~\cite{tanna2026finetuning} and tabular ICL architectures~\cite{bouadi2026orionbix} have begun adapting TFMs to such settings; our work is complementary, focused on what happens \emph{after} a strong TFM is available and the bottleneck is deployment. The TabTune library~\cite{tanna2025tabtune} provides a unified evaluation surface for accuracy, calibration, and fairness, and is the basis for our experimental harness.

\section{Methodology}
\label{sec:methodology}

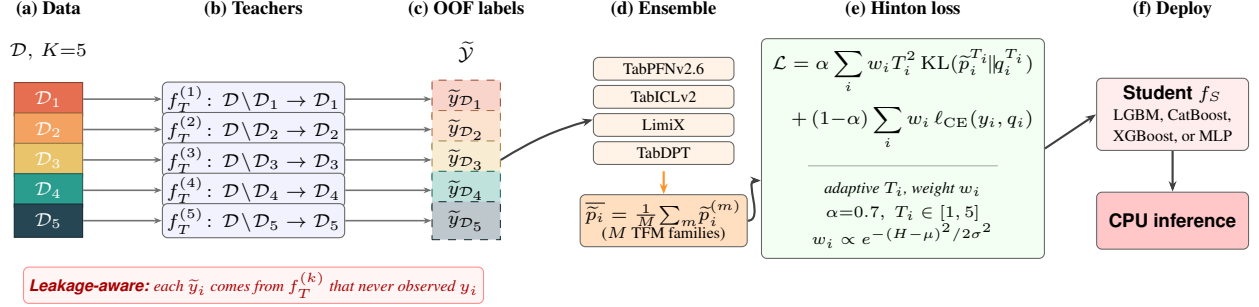
\begin{figure}[t]
\centering
\resizebox{\textwidth}{!}{%
\begin{tikzpicture}[
  every node/.style={font=\scriptsize},
  box/.style={rectangle, draw=black!60, rounded corners=2pt, align=center, font=\scriptsize, inner sep=2.5pt},
  fold/.style={rectangle, minimum width=0.9cm, minimum height=0.4cm, draw=black!40, font=\scriptsize},
  arr/.style={-{Stealth[length=4pt]}, thick, draw=black!75},
  thinarr/.style={-{Stealth[length=3pt]}, semithick, draw=black!55},
  stagelabel/.style={font=\bfseries\scriptsize, anchor=south},
]

\definecolor{f1}{RGB}{231,111,81}
\definecolor{f2}{RGB}{244,162,97}
\definecolor{f3}{RGB}{233,196,106}
\definecolor{f4}{RGB}{42,157,143}
\definecolor{f5}{RGB}{38,70,83}

\node[stagelabel] at (0.6, 1.7) {(a) Data};
\node[font=\scriptsize] at (0.6, 1.38) {$\mathcal{D},\, K{=}5$};
\foreach \i/\c in {1/f1, 2/f2, 3/f3, 4/f4, 5/f5} {
    \node[fold, fill=\c, draw=\c!60!black] at (0.6, 1.15 - \i*0.4) (D\i) {\textcolor{white}{\bfseries\scriptsize $\mathcal{D}_\i$}};
}

\node[stagelabel] at (3.3, 1.7) {(b) Teachers};
\foreach \k in {1, 2, 3, 4, 5} {
    \node[box, fill=blue!5, minimum width=2.4cm, minimum height=0.4cm, align=center, inner sep=1.5pt] at (3.3, 1.15 - \k*0.4) (T\k) {%
        \scriptsize$f_T^{(\k)}\!:\,\mathcal{D}\!\setminus\!\mathcal{D}_\k \to \mathcal{D}_\k$%
    };
    \draw[thinarr] (D\k.east) -- (T\k.west);
}

\node[stagelabel] at (6.1, 1.7) {(c) OOF labels};
\node[font=\scriptsize] at (6.1, 1.38) {$\widetilde{\mathcal{Y}}$};
\foreach \i/\c in {1/f1, 2/f2, 3/f3, 4/f4, 5/f5} {
    \node[fold, fill=\c!30, draw=\c!70!black, dashed] at (6.1, 1.15 - \i*0.4) (O\i) {$\widetilde{y}_{\mathcal{D}_\i}$};
    \draw[thinarr] (T\i.east) -- (O\i.west);
}

\node[draw=red!45, fill=red!4, rounded corners=2pt, font=\tiny\itshape, inner sep=2pt, align=center, text=red!65!black]
    at (3.3, -1.7) {\textbf{Leakage-aware:} each $\widetilde{y}_i$ comes from $f_T^{(k)}$ that never observed $y_i$};

\node[stagelabel] at (8.7, 1.7) {(d) Ensemble};

\node[box, fill=orange!10, minimum width=1.85cm, minimum height=0.30cm, font=\tiny, inner sep=1.5pt] at (8.7, 1.15) (E1) {TabPFNv2.6};
\node[box, fill=orange!10, minimum width=1.85cm, minimum height=0.30cm, font=\tiny, inner sep=1.5pt] at (8.7, 0.78) (E2) {TabICLv2};
\node[box, fill=orange!10, minimum width=1.85cm, minimum height=0.30cm, font=\tiny, inner sep=1.5pt] at (8.7, 0.41) (E3) {LimiX};
\node[box, fill=orange!10, minimum width=1.85cm, minimum height=0.30cm, font=\tiny, inner sep=1.5pt] at (8.7, 0.04) (E4) {TabDPT};

\coordinate (mergeL) at (7.95, -0.20);
\coordinate (mergeR) at (9.45, -0.20);
\coordinate (mergeC) at (8.7, -0.20);


\node[box, fill=orange!25, minimum width=2.05cm, minimum height=0.65cm, font=\scriptsize, align=center, inner sep=2pt] at (8.7, -0.85) (ENS)
    {$\overline{\widetilde{p}_i}=\tfrac{1}{M}\!\sum_m\! \widetilde{p}^{(m)}_i$\\[-1pt]\tiny ($M$ TFM families)};

\draw[arr, draw=orange!85] (mergeC) -- (ENS.north);

\draw[arr] (O3.east) to[bend left=8] ([yshift=2pt]E3.west);

\node[stagelabel] at (11.85, 1.7) {(e) Hinton loss};

\node[box, fill=green!5, minimum width=3.0cm, minimum height=2.05cm, align=center, font=\scriptsize, inner sep=4pt] at (11.85, 0.10) (LOSS) {%
$\mathcal{L} = \alpha \displaystyle\sum_i w_i T_i^2\, \mathrm{KL}(\widetilde{p}_i^{T_i} \!\|\! q_i^{T_i})$\\[3pt]
$\quad+\, (1{-}\alpha) \displaystyle\sum_i w_i\, \ell_{\mathrm{CE}}(y_i, q_i)$\\[5pt]
{\color{black!35}\rule{2.5cm}{0.3pt}}\\[2pt]
{\tiny\itshape adaptive $T_i$, weight $w_i$}\\[1pt]
{\tiny $\alpha{=}0.7,\; T_i \in [1, 5]$}\\
{\tiny $w_i \propto e^{-(H-\mu)^2/2\sigma^2}$}%
};

\draw[arr] (ENS.east) to[out=0,in=180] ([yshift=-12pt]LOSS.west);

\node[stagelabel] at (15.4, 1.7) {(f) Deploy};

\node[box, fill=red!7, minimum width=2.0cm, align=center, minimum height=0.95cm] at (15.4, 0.55) (S) {\textbf{Student $f_S$}\\\tiny LGBM, CatBoost,\\\tiny XGBoost, or MLP};

\node[box, fill=red!22, minimum width=2.0cm, align=center, minimum height=0.75cm, font=\scriptsize] at (15.4, -0.85) (DEPLOY) {%
\textbf{CPU inference}%
};

\draw[arr] (LOSS.east) -- (S.west);
\draw[arr] (S.south) -- (DEPLOY.north);

\end{tikzpicture}%
}
\caption{\small \textbf{Leakage-aware out-of-fold distillation pipeline.} (a)~The training set $\mathcal{D}$ is partitioned into $K{=}5$ stratified folds $\mathcal{D}_1,\ldots,\mathcal{D}_5$. (b)~For each fold $k$, a teacher TFM $f_T^{(k)}$ is conditioned on $\mathcal{D}\!\setminus\!\mathcal{D}_k$ and predicts only on $\mathcal{D}_k$. (c)~The fold-wise predictions are concatenated into the out-of-fold soft-label set $\widetilde{\mathcal{Y}}$, so no soft target $\widetilde{y}_i$ is generated by a teacher that conditioned on $y_i$. (d)~In the multi-teacher setting, OOF labels from $M$ different TFM families (TabPFNv2.6, TabICLv2, LimiX, TabDPT, $\ldots$) are averaged with equal weights. (e)~The student is trained with a Hinton mixed loss combining temperature-scaled KL on the soft targets and cross-entropy on the hard labels, with per-sample adaptive temperature $T_i\!\in\![1,5]$ and confidence weight $w_i$ peaking on moderate-entropy teacher predictions. (f)~The trained student is deployed on CPU}\textsc{10pt}
\label{fig:pipeline}
\end{figure}

\paragraph{Formalization.} In this work, we consider a set of tabular datasets $\mathcal{D} = \{(\mathbf{x}_i, y_i)\}_{i=1}^n$ where $\mathbf{x}_i \in \mathbb{R}^{l_i \times c_i}$ and $y_i \in \mathbb{R}^{l_i}$; $l_i$ is the sequence length and $c_i$ is the number of features for the $i$-th dataset. We also consider a Tabular Foundation Model (TFM) as a teacher $f_T$ that predicts the labels using a function $f_T(\cdot): \mathbb{R}^{l \times c} \to \mathbb{R}^{l}$.

Given a set of datasets $\mathcal{D}$ and a TFM $f_T$, we aim to find a student $f_S: \mathbb{R}^{l \times c} \to \mathbb{R}^{l}$ that approximates $f_T$ without duplicating the teacher's weights, gradients, or hidden states.

\paragraph{Distillation approach.} To transfer the predictive behavior of $f_T$ to $f_S$, we employ knowledge distillation~\cite{hinton2015distilling}, training $f_S$ to match the outputs of $f_T$ rather than hard ground-truth labels. This choice is motivated by recent empirical works showing that distillation leads to stronger student generalization~\cite{tang2020understanding, zhou2021rethinking}.

We define the student objective as the Hinton mixed loss:
\begin{equation}
\mathcal{L} = \alpha \sum_i w_i T_i^2 \, \mathrm{KL}(\hat{p}_i^{T_i} \,\|\, q_i^{T_i}) + (1-\alpha) \sum_i w_i \, \ell_{\mathrm{CE}}(y_i, q_i),
\end{equation}
with $\alpha{=}0.7$. For the three tree-based students, the soft loss reduces to a per-sample weighted MSE on soft-label logits.

\paragraph{Adaptive temperature.} Per-sample $T_i \in [T_{\min}, T_{\max}]$ scales each $\tilde p_i$ as a function of teacher entropy $H(\tilde p_i)$: confident predictions get $T_i \approx T_{\min}{=}1$, ambiguous ones get $T_i \approx T_{\max}{=}5$, following~\cite{guo2017calibration}. Confidence weight $w_i = \exp(-(H(\tilde p_i) - \mu)^2 / 2\sigma^2)$ peaks on moderate-entropy predictions ($(\mu,\sigma){=}(0.7, 0.2)$).

\paragraph{Teacher soft labels.} In our work, all teachers are TFMs. While all these models rely on in-context learning for predictions, the teachers condition on the training set as part of their input context. More formally, we denote the teacher as $f_T(\cdot \mid C)$ where $C = \{x_{i}, y_{i}\}$ is the context set. Predictions over the samples $x_i$ are therefore produced by a teacher that has already observed their targets $y_i$, yielding soft labels that are overconfident compared to the predicted distribution on unseen data.

To mitigate this, we apply $k{=}5$ stratified cross-validation, fitting each teacher $f_T^{(k)}$ on $\mathcal{D}\!\setminus\!\mathcal{D}_k$ and predicting only over $\mathcal{D}_k$, ensuring that no sample's soft target is produced by a $f_T$ that conditioned on it.

\paragraph{Multi-teacher averaging.} When more than one TFM family is available, we run the OOF labeling pipeline independently for each teacher and then average the resulting soft labels with equal weights: $\overline{\widetilde{p}_i} = \frac{1}{M}\sum_{m=1}^{M} \widetilde{p}_i^{(m)}$, with the multi-teacher case shown in \Cref{fig:pipeline}(d). We did not learn ensemble weights or stack the teachers in a second-stage model. The goal here is the cleanest possible baseline for the question \emph{does the student get more accurate when you give it more teachers?} \Cref{sec:experiments} reports results for several subsets of $\{$TabPFNv2.6, TabICLv2, LimiX, TabDPT, Orion-MSP$\}$.

We treat OOF prediction as a correctness requirement rather than a tunable hyperparameter: every standard distillation pipeline assumes labels generated on points the teacher has not seen. The choice of $K{=}5$ follows the stacking convention. Pilot runs at $K\in\{3, 10\}$ moved AUC by less than $0.005$ at $0.6\times$ and $2\times$ the teacher cost, respectively, so we did not tune this further.

\section{Experiments}
\label{sec:experiments}

\subsection{Setup}
\paragraph{Datasets.} We use $19$ health datasets, of which $16$ are binary and $3$ multiclass, covering cardiology, oncology, nephrology, dermatology, and critical care, with sample sizes from $299$ to $9105$ (bigger datasets were sampled from) and feature counts from $4$ to $40$. Full details are in \Cref{tab:datasets}.

\paragraph{Teacher TFMs.} We consider $6$ TFMs: TabPFNv2.5~\cite{hollmann2025tabpfn}, TabPFNv2.6~\cite{grinsztajn2025tabpfn25}, TabICLv2~\cite{qu2025tabiclv2}, TabDPT~\cite{ma2025tabdpt}, LimiX~\cite{zhang2025limix}, and Orion-MSP v1.5~\cite{bouadi2025orionmsp}, each with its built-in missing-value handler. For multi-teacher ($M{>}1$) settings, we average per-fold soft labels with equal weights.

\paragraph{Students.} We use LightGBM~\cite{ke2017lightgbm}, CatBoost~\cite{prokhorenkova2018catboost}, XGBoost~\cite{chen2016xgboost} (300 trees, depth 6, patience-30 early stopping) and an MLP (embedding $\min(8d,128)$, cosine LR with warmup, label smoothing 0.05, SWA on final 20\% of training, collapse-detector restart at higher dropout).

\paragraph{Baselines.} We use LogisticRegression, XGBoost, and LightGBM with a fixed configuration on zero-imputed inputs. These baselines reflect how each model class is typically used out of the box.

\paragraph{Metrics.} We report (i) average AUC; (ii) retention, defined as $\frac{\mathrm{AUC}(\text{Student})}{\mathrm{AUC}(\text{Teacher})}\times 100$; (iii) model latency and memory consumption; (iv) calibration (ECE, Brier) and fairness (DP-diff, EO-diff). All reported results are aggregated over $5$ simulations. All models and runs use identical pre-processing and evaluation as implemented in TabTune~\cite{tanna2025tabtune}.

\subsection{Empirical Results}
\paragraph{Accuracy.} \Cref{tab:accuracy} shows the average AUC and retention on LGBM and MLP on the $16$ binary datasets. Results for the other student models are in \Cref{tab:accuracy_all}, and the corresponding multiclass numbers are in \Cref{tab:accuracy_multi_class}.

\begin{table}[t]
\caption{Avg AUC and retention on LGBM and MLP over the binary datasets.}
\label{tab:accuracy}
\centering
\resizebox{0.65\columnwidth}{!}{%
\begin{tabular}{llcc}
\toprule
Type & Model & AUC & Ret. \\
\midrule
\multirow{3}{*}{Baseline}
 & LogisticRegression & \best{.857} $\pm$ .108 & -- \\
 & XGBoost & .854 $\pm$ .133 & -- \\
 & LightGBM & .853 $\pm$ .134 & -- \\
\midrule
\multirow{6}{*}{Teacher}
 & LimiX & \best{.875} $\pm$ .119 & -- \\
 & TabICLv2 & .873 $\pm$ .124 & -- \\
 & TabPFNv2.6 & .870 $\pm$ .125 & -- \\
 & TabDPT & .868 $\pm$ .126 & -- \\
 & TabPFNv2.5 & .868 $\pm$ .125 & -- \\
 & Orion-MSP v1.5 & .864 $\pm$ .124 & -- \\
\midrule
\multirow{6}{*}{\shortstack[l]{Distilled\\(LGBM)}}
 & LimiX$\to$LGBM & \best{.865} $\pm$ .119 & 98.8\% \\
 & TabICLv2$\to$LGBM & .863 $\pm$ .122 & 98.9\% \\
 & TabPFNv2.6$\to$LGBM & .862 $\pm$ .123 & 99.1\% \\
 & TabPFN$\to$LGBM & .862 $\pm$ .124 & 99.4\% \\
 & TabDPT$\to$LGBM & .861 $\pm$ .123 & 99.1\% \\
 & Orion-MSP$\to$LGBM & .860 $\pm$ .119 & 99.6\% \\
\midrule
\multirow{4}{*}{\shortstack[l]{Distilled\\(LGBM,\\multi-teacher)}}
 & {[}PFN+OrionMSP+Limix{]}$\to$LGBM & \best{.864} $\pm$ .121 & -- \\
 & {[}PFN+Limix{]}$\to$LGBM & \best{.864} $\pm$ .121 & -- \\
 & {[}PFN+ICL+Limix{]}$\to$LGBM & .863 $\pm$ .123 & -- \\
 & {[}PFN+ICL+Limix+TabDPT{]}$\to$LGBM & .862 $\pm$ .123 & -- \\
\midrule
\multirow{6}{*}{\shortstack[l]{Distilled\\(MLP)}}
 & LimiX$\to$MLP & \best{.795} $\pm$ .173 & 90.8\% \\
 & TabPFN$\to$MLP & .787 $\pm$ .184 & 90.7\% \\
 & TabPFNv2.6$\to$MLP & .782 $\pm$ .189 & 89.9\% \\
 & TabICLv2$\to$MLP & .782 $\pm$ .190 & 89.6\% \\
 & TabDPT$\to$MLP & .781 $\pm$ .190 & 89.9\% \\
 & Orion-MSP$\to$MLP & .775 $\pm$ .182 & 89.7\% \\
\midrule
\multirow{4}{*}{\shortstack[l]{Distilled\\(MLP,\\multi-teacher)}}
 & {[}PFN+OrionMSP+Limix{]}$\to$MLP & \best{.790} $\pm$ .181 & -- \\
 & {[}PFN+Limix{]}$\to$MLP & .787 $\pm$ .180 & -- \\
 & {[}PFN+ICL+Limix+TabDPT{]}$\to$MLP & .787 $\pm$ .182 & -- \\
 & {[}PFN+ICL+Limix{]}$\to$MLP & .785 $\pm$ .182 & -- \\
\bottomrule
\end{tabular}}
\end{table}

Distillation is strongly \textit{student-dependent}. Tree-based students benefit most: LightGBM retains $98.8$--$99.6\%$ of teacher AUC and improves over the default LightGBM baseline, while \Cref{tab:accuracy_all} shows that CatBoost and XGBoost can reach or exceed $100\%$ retention. This suggests that distilled students can sometimes outperform their teachers, consistent with distillation acting as a \textbf{regularizer through softened targets rather than only as compression}. In contrast, MLP students retain only $89.6$--$90.8\%$ and remain below the default baselines. The tree--MLP retention gap is consistent with what tabular ML has long observed: tree splits handle heterogeneous feature scales without preprocessing, while MLPs need careful regularization to compete. On our small-cohort benchmark, the MLP student is asked to regularize and distil at the same time; \Cref{app:limits} discusses the specific failure modes we observed. Multi-teacher distillation also gives limited gains: except for CatBoost, the best multi-teacher student does not outperform the best single-teacher student, suggesting that uniform averaging can add noise when teachers disagree~\cite{du2020agree}.

\paragraph{Latency and memory.} \Cref{tab:latency} reports latency on 6 datasets; full results are in \Cref{tab:latency_all}. Students run on CPU, while TFM teachers run on GPU. Distilled LGBM and MLP students require about $7$\,ms and $3.8$\,ms, respectively, versus $187.4$\,ms for the fastest teacher, yielding roughly $26\times$ and $49\times$ lower latency while staying under $1$\,MB. This makes distilled students suitable for CPU-only, high-throughput healthcare deployment: batch retrospective scoring across millions of records, or ICU streaming inference where every millisecond counts, become straightforward at $49$K--$80$K predictions per second per CPU core.

\begin{table}[t]
\caption{Inference latency and throughput across datasets}
\label{tab:latency}
\centering
\resizebox{0.65\columnwidth}{!}{%
\begin{tabular}{lrrr}
\toprule
Model & Latency (ms) & Throughput (predictions/s) \\
\midrule
\multicolumn{3}{l}{\textit{TFM teachers (GPU)}} \\
\midrule
TabICLv2     & \best{187.4}   & \best{3{,}221/s} \\
LimiX        & 433.8   & 1{,}450/s \\
TabPFNv2.6   & 564.9   & 1{,}099/s \\
Orion-MSP v1.5 & 1{,}911.9 & 323/s \\
TabDPT       & 4{,}010.0 & 600/s \\
\midrule
\multicolumn{3}{l}{\textit{Distilled MLP students (CPU)}} \\
\midrule
LimiX$\to$MLP                                & \best{3.7} & \best{172K/s} \\
TabPFNv2.6$\to$MLP                           & 3.8 & 170K/s \\
TabICLv2$\to$MLP                             & 3.8 & 169K/s \\
{[}PFNv2.6+Limix{]}$\to$MLP                  & 3.8 & 166K/s \\
\midrule
\multicolumn{3}{l}{\textit{Distilled LGBM students (CPU)}} \\
\midrule
LimiX$\to$LGBM                               & \best{7.0} & 87K/s \\
{[}PFNv2.6+ICL+Limix{]}$\to$LGBM             & 7.1 & 92K/s \\
TabPFNv2.6$\to$LGBM                          & 7.3 & 82K/s \\
{[}PFNv2.6+Limix{]}$\to$LGBM                 & 7.7 & \best{98K/s} \\
\bottomrule
\end{tabular}}
\end{table}

\paragraph{Calibration.} \Cref{tab:calfair} reports ECE and Brier scores on 12 binary datasets, with and without global temperature scaling (TS)~\cite{guo2017calibration,platt1999probabilistic}. Distilled LGBM students preserve much of the teachers' calibration, with ECE around $.058 $-$ .063$, substantially better than default LightGBM and XGBoost. TS provides only small additional gains for LGBM students, suggesting that they are already reasonably calibrated. In contrast, MLP students are poorly calibrated before TS, with ECE above $.12$, but improve markedly after scaling. Thus, calibration quality is strongly student-dependent: distilled trees are reliable without much post-hoc correction, whereas MLP students require calibration.

Two further observations are worth noting. First, the default GBDT baselines (XGBoost, LightGBM) show the largest absolute ECE drop from temperature scaling ($0.090\to0.069$), consistent with the well-known tendency of tree models to be overconfident at the leaves~\cite{niculescu2005predicting}. Any deployment that uses raw GBDT scores as risk probabilities will therefore need a calibration step. Second, the MLP miscalibration is dominated by overconfidence on the larger cohorts (\texttt{support2}, \texttt{sick}); on reliability diagrams the high-confidence bins consistently overshoot, which TS rescales globally but does not fix per-region.

\begin{table}[t]
\caption{Avg ECE and Brier on $12$ binary datasets w/o global temperature scaling (TS).}
\label{tab:calfair}
\centering
\resizebox{0.65\columnwidth}{!}{%
\begin{tabular}{lcccc}
\toprule
Model & ECE $\downarrow$ & ECE+TS $\downarrow$ & Brier $\downarrow$ & Brier+TS $\downarrow$ \\
\midrule
\multicolumn{5}{l}{\textit{Baselines}} \\
\midrule
LogisticRegression                    & \best{.062} & \best{.063} & \best{.101} & \best{.100} \\
LightGBM                              & .090 & .069 & .116 & .106 \\
XGBoost                               & .092 & .069 & .116 & .107 \\
\midrule
\multicolumn{5}{l}{\textit{Teachers}} \\
\midrule
TabPFNv2.6                            & \best{.053} & \best{.053} & .095 & .094 \\
TabDPT                                & .056 & \best{.053} & .094 & \best{.093} \\
LimiX                                 & .056 & .055 & \best{.093} & \best{.093} \\
TabICLv2                              & .056 & \best{.053} & \best{.093} & \best{.093} \\
\midrule
\multicolumn{5}{l}{\textit{Distilled LGBM students (single-teacher)}} \\
\midrule
TabDPT$\to$LGBM                       & \best{.060} & \best{.057} & .100 & .099 \\
TabPFNv2.6$\to$LGBM                   & .061 & .060 & \best{.098} & \best{.098} \\
\midrule
\multicolumn{5}{l}{\textit{Distilled LGBM students (multi-teacher)}} \\
\midrule
{[}PFNv2.6+Limix{]}$\to$LGBM          & \best{.058} & .061 & \best{.099} & .099 \\
{[}PFNv2.6+OrionMSP+Limix{]}$\to$LGBM & .063 & \best{.059} & \best{.099} & \best{.098} \\
\midrule
\multicolumn{5}{l}{\textit{Distilled MLP students (single-teacher)}} \\
\midrule
TabPFNv2.6$\to$MLP                    & \best{.123} & .074 & \best{.148} & \best{.133} \\
TabDPT$\to$MLP                        & .124 & \best{.072} & .149 & \best{.133} \\
\midrule
\multicolumn{5}{l}{\textit{Distilled MLP students (multi-teacher)}} \\
\midrule
{[}PFNv2.6+Limix{]}$\to$MLP           & \best{.124} & .076 & .150 & \best{.133} \\
{[}PFNv2.6+ICL+Limix+TabDPT{]}$\to$MLP & .125 & \best{.075} & \best{.149} & \best{.133} \\
\bottomrule
\end{tabular}}
\end{table}

\paragraph{Fairness.} \Cref{tab:fairness} reports DP- and EO-difference across 8 datasets and 4 sensitive attributes. Negative $\Delta$ values mean the student has a smaller fairness gap than its teacher. Distilled LGBM students reduce DP gaps and EO gaps in most cases, with average $\Delta$DP $=-0.013$ and $\Delta$EO $=-0.014$, consistent with soft-label smoothing in a smaller hypothesis class. MLP students reduce DP more strongly ($\Delta$DP $=-0.034$), but often increase EO gaps ($\Delta$EO $=+0.041$). Thus, LGBM provides a more stable fairness tradeoff, while MLP can improve parity in prediction rates at the cost of subgroup error disparities.

\begin{table}[t]
\caption{Group fairness across $8$ datasets and $4$ attributes. In multi-teacher settings, $\Delta$ are computed with TabPFNv2.6.}
\label{tab:fairness}
\centering
\resizebox{0.65\columnwidth}{!}{%
\begin{tabular}{lcccc}
\toprule
Model & DP-diff $\downarrow$ & EO-diff $\downarrow$ & $\Delta$DP & $\Delta$EO \\
\midrule
\multicolumn{5}{l}{\textit{Baselines}} \\
\midrule
LogisticRegression                      & \best{.131} & .183 & -- & -- \\
XGBoost                                 & .145 & \best{.165} & -- & -- \\
LightGBM                                & .145 & .167 & -- & -- \\
\midrule
\multicolumn{5}{l}{\textit{Teachers}} \\
\midrule
LimiX                                   & \best{.134} & \best{.144} & -- & -- \\
TabDPT                                  & .143 & .156 & -- & -- \\
TabPFNv2.6                              & .145 & \best{.144} & -- & -- \\
TabICLv2                                & .154 & .166 & -- & -- \\
Orion-MSP v1.5                          & .153 & .202 & -- & -- \\
\midrule
\multicolumn{5}{l}{\textit{Distilled LGBM students (single-teacher)}} \\
\midrule
LimiX$\to$LGBM                          & \best{.130} & \best{.125} & $-.005$ & $-.018$ \\
OrionMSPv1.5$\to$LGBM                   & .132 & .173 & $-.021$ & $-.029$ \\
TabPFNv2.6$\to$LGBM                     & .131 & \best{.125} & $-.014$ & $-.019$ \\
TabDPT$\to$LGBM                         & .136 & .171 & $-.007$ & $+.015$ \\
TabICLv2$\to$LGBM                       & .139 & .136 & $-.015$ & $-.031$ \\
\midrule
\multicolumn{5}{l}{\textit{Distilled LGBM students (multi-teacher)}} \\
\midrule
{[}PFNv2.6+OrionMSP+Limix{]}$\to$LGBM   & \best{.129} & .130 & $-.016$ & $-.014$ \\
{[}PFNv2.6+Limix{]}$\to$LGBM            & .133 & .126 & $-.012$ & $-.017$ \\
{[}PFNv2.6+ICL+Limix{]}$\to$LGBM        & .135 & \best{.120} & $-.010$ & $-.024$ \\
\midrule
\multicolumn{5}{l}{\textit{Distilled MLP students (single-teacher)}} \\
\midrule
OrionMSPv1.5$\to$MLP                    & \best{.104} & .216 & $-.049$ & $+.014$ \\
TabICLv2$\to$MLP                        & .106 & .165 & $-.048$ & $-.001$ \\
LimiX$\to$MLP                           & .108 & .187 & $-.026$ & $+.043$ \\
TabPFNv2.6$\to$MLP                      & .108 & \best{.162} & $-.036$ & $+.019$ \\
TabDPT$\to$MLP                          & .117 & .178 & $-.026$ & $+.022$ \\
\midrule
\multicolumn{5}{l}{\textit{Distilled MLP students (multi-teacher)}} \\
\midrule
{[}PFNv2.6+OrionMSP+Limix{]}$\to$MLP    & \best{.111} & \best{.193} & $-.034$ & $+.049$ \\
{[}PFNv2.6+ICL+Limix{]}$\to$MLP         & .114 & .207 & $-.030$ & $+.063$ \\
{[}PFNv2.6+Limix{]}$\to$MLP             & .116 & .219 & $-.029$ & $+.075$ \\
\bottomrule
\end{tabular}}
\end{table}

The pattern is consistent with a smaller hypothesis class plus soft-label smoothing acting as a fairness regularizer for the LGBM student. The MLP shifts in the opposite direction on EO: smaller prediction-rate gaps, but larger subgroup error disparities on \texttt{pima\_diabetes}, \texttt{sick}, and a few of the larger cohorts. Practitioners should treat the EO regression as a real concern and audit it on the actual sensitive attributes of their deployment, not extrapolate from the table alone.

\paragraph{Component ablation.} \Cref{tab:ablation} ablates MLP distillation with TabPFNv2.6 on 5 datasets. Hard-label training outperforms the full setup by $0.034$ AUC ($p{=}0.004$), while removing adaptive temperature, confidence weighting, or augmentation changes AUC by at most $0.02$. Thus, for MLPs, the soft-label machinery adds little on this benchmark; we leave tree-student ablations to future work.

\begin{table}[h!]
\caption{Ablation on the MLP student with TabPFNv2.6 teacher on $5$ datasets. $\Delta$ is the paired difference vs.\ \emph{full}; $p$ from a Wilcoxon signed-rank test on per-(dataset, simulation) deltas.}
\label{tab:ablation}
\centering
\footnotesize
\begin{tabular}{lccc}
\toprule
Configuration & AUC & $\Delta$ vs.\ full & $p$ \\
\midrule
Full (all components) & .829 & -- & -- \\
No adaptive temperature & .809 & $-.020$ & .49 \\
No confidence weighting & .828 & $-.001$ & .93 \\
No augmentation & .818 & $-.011$ & .34 \\
Hard labels only ($\alpha{=}0$) & \best{.863} & $+.034$ & .004 \\
Soft labels only ($\alpha{=}1$) & .814 & $-.014$ & .54 \\
Low temperature ($T{=}1$) & .855 & $+.027$ & .06 \\
High temperature ($T{=}5$) & .810 & $-.018$ & .19 \\
\bottomrule
\end{tabular}
\end{table}

\section{Discussion and Limitations}
\label{app:limits}

In this section, we discuss the main limitations of our work and a few design decisions worth flagging for practitioners.

\paragraph{MLP instability on small low-dimensional inputs.} The MLP student is unstable on small low-dimensional inputs. On three datasets with under $750$ samples and $4$--$8$ features (\texttt{blood\_transfusion}, \texttt{indian\_liver\_patient}, \texttt{pima\_diabetes}), it showed high seed-to-seed variance and occasionally collapsed to near-random predictions. The library has a collapse detector that restarts training with higher dropout, but the underlying issue is that a residual MLP is overparameterised for these inputs. The LGBM student does not have this problem. Recommendation: use LGBM by default and consider an MLP student only when inference latency under $2$\,ms is a hard requirement.

\paragraph{Multiclass distillation gap.} TabPFNv2.5 scores $0.985$ AUC on the kidney disease dataset ($3$ classes, $400$ rows), while its distilled TabPFN$\to$LGBM version scores only $0.749$. The reason is that the soft-label distillation breaks for problems in the $3+$ class due to the LGBM regressor formulation we use. A multinomial-loss formulation would be the proper fix; until that lands, multiclass distillation should be treated as experimental.

\paragraph{Why uniform multi-teacher averaging did not help.} \Cref{tab:accuracy} shows the best multi-teacher LGBM student tied, rather than beat, the best single-teacher student at three decimal places. Two factors are likely at work. The TFM teachers are themselves close in AUC ($0.864$--$0.875$), so the headroom for ensemble gain on these datasets was small to begin with. And equal-weight averaging treats teacher disagreement as additional information even when the disagreement is closer to noise; \cite{du2020agree} formalises this for distillation. A learned or accuracy-weighted scheme could plausibly close the gap, especially in regimes where one teacher is meaningfully stronger than the others. We leave that for follow-up work.

\paragraph{Hard-label ablation.} \Cref{tab:ablation} shows that hard-label training ($\alpha{=}0$) outperforms the full pipeline ($\alpha{=}0.7$) by $0.034$ AUC ($p{=}0.004$, Wilcoxon) on the MLP. This does not affect the need for out-of-fold labeling: for ICL teachers, scoring examples that appear in the context can produce overconfident soft targets and leakage. Rather, the result suggests that, on small relatively clean datasets, the added soft-label components do not improve MLP training. Soft targets may be more useful in noisier or higher-dimensional clinical settings, where hard labels are less reliable. The matching LGBM ablation is left to future work; on partial LGBM ablation runs we have completed, the components again contribute less than the per-dataset variance.

\section{Conclusion}
\label{sec:conclusion}

In this work, we argued that TFMs predict well on small health datasets but are costly to deploy in inference-constrained health settings. We proposed a methodology around knowledge distillation to transfer predictive behavior from TFMs to simpler standard models, and ran extensive experiments on $19$ healthcare datasets. Our results show that distilled models run, on CPU, at least $26\times$ faster than the best TFMs on GPU. While knowledge transfer recovers at least $90$\% of TFM accuracy, it also preserves calibration and fairness. As a practical recommendation: use an LGBM student by default, and consider an MLP student only when sub-$2$\,ms latency is a hard requirement.

More broadly, this paper argues that TFM inference cost reflects how TFMs do inference (context-conditioned GPU forward passes) rather than something intrinsic to their accuracy. Out-of-fold distillation moves that cost off the deployment surface without losing the accuracy.

\section{Impact and Ethical Considerations}
\label{sec:impact}

\paragraph{Lowering the deployment bar.} The practical effect of this work is that TFM-quality predictions on small clinical datasets no longer require a GPU. A $7$\,ms LightGBM student that fits in under $1$\,MB can run inside hospital systems that today cannot host a TabPFN forward pass: CPU-only nodes, air-gapped environments, registry-scale batch jobs, embedded scoring at the point of care. This widens the set of settings where recent TFM accuracy is reachable, particularly in resource-limited deployments or those with data-residency constraints.

\paragraph{Bias inheritance.} A distilled student inherits whatever bias the teacher carries. On the 8-dataset, 4-attribute fairness benchmark we report, distilled LGBM students stayed close to teacher DP- and EO-difference, with average $\Delta$DP of $-0.013$ and average $\Delta$EO of $-0.014$. MLP students cut DP-difference more strongly but increased EO-difference on several datasets, meaning prediction rates equalised at the cost of subgroup error rates. Any deployment should audit the specific student-teacher pair on the actual sensitive attributes that matter for the use case. Generalising from our table to a different cohort or attribute is not safe.

\paragraph{Calibration and decision support.} A clinical model whose probabilities are off cannot be plugged into a threshold-based decision rule without a calibration step. Distilled LightGBM students preserve teacher calibration without much post-hoc correction; MLP students need temperature scaling before deployment. Practitioners should not skip this step.

\clearpage

\bibliographystyle{unsrt}
\bibliography{references}

\newpage
\appendix

\section{Implementation details}
\label{app:impl}

Defaults: $K{=}5$ folds, $T_{\min}{=}1$, $T_{\max}{=}5$, $\alpha{=}0.7$, confidence-weight $(\mu, \sigma) = (0.7, 0.2)$. LGBM students use 300 estimators with patience-30 early stopping. MLP students use embedding dimension $\min(8d, 128)$, hidden widths scaled to dataset size, warmup-plus-cosine LR, label smoothing 0.05, and SWA on the final 20\% of training. A collapse detector monitors prediction entropy and restarts with higher dropout if degenerate predictions appear. Teachers run inference-only with no fine-tuning. Latency is measured on an Intel Xeon 8358 CPU for students and an NVIDIA A100 GPU for teachers, averaged over 50 runs (students) and 20 runs (teachers).

\section{Dataset details}
\label{app:datasets}

\begin{table}[h]
\caption{The 19 health datasets used. Modality codes: C = cardiology, O = oncology, N = nephrology, D = dermatology, E = endocrinology, ICU = intensive care, H = hepatology, OG = obstetrics-gynaecology, W = women's health.}
\label{tab:datasets}
\centering
\resizebox{0.9\textwidth}{!}{%
\begin{tabular}{llrrrl}
\toprule
Dataset & Modality & Samples & Features & Classes & Use \\
\midrule
heart\_cleveland          & C   & 303     & 13 & 2 & accuracy, calibration, fairness \\
heart\_failure\_clinical  & C   & 299     & 12 & 2 & accuracy, calibration, fairness \\
south\_african\_heart     & C   & 462     &  9 & 2 & accuracy, calibration \\
breast\_cancer\_wisconsin & O   & 569     & 30 & 2 & accuracy, calibration \\
wdbc                       & O   & 569     & 30 & 2 & accuracy, calibration \\
breast\_w                  & O   & 699     &  9 & 2 & accuracy, calibration \\
indian\_liver\_patient    & H   & 583     & 10 & 2 & accuracy, calibration \\
blood\_transfusion         & --- & 748     &  4 & 2 & accuracy, calibration \\
pima\_diabetes             & E   & 768     &  8 & 2 & accuracy, calibration, fairness \\
cervical\_cancer           & W   & 858     & 35 & 2 & accuracy  \\
mammographic\_mass         & O   & 961     &  5 & 2 & accuracy, calibration \\
sick                       & E   & 3{,}772 & 27 & 2 & accuracy, calibration, fairness \\
support2                   & ICU & 9{,}105 & 21 & 2 & accuracy, calibration, fairness \\
framingham                 & C   & 4{,}238 & 15 & 2 & accuracy, fairness (sex) \\
diabetes\_130\_us          & E   & 101{,}766 & 47 & 2 & accuracy, fairness (race) \\
mimic\_iii\_mortality      & ICU & 58{,}976 & 23 & 2 & accuracy, fairness (gender) \\
chronic\_kidney\_disease   & N   & 400     & 24 & 3 & accuracy  \\
dermatology                & D   & 366     & 34 & 6 & accuracy \\
analcatdata\_dmft          & OG  & 797     &  4 & 6 & accuracy \\
\bottomrule
\end{tabular}}
\end{table}

\Cref{tab:datasets} lists all 19 datasets with sample size, feature count, class count, and the role each plays in the evaluation. Three of the binary cohorts are versions of the Wisconsin breast-cancer dataset (\texttt{breast\_cancer\_wisconsin}, \texttt{wdbc}, \texttt{breast\_w}); we kept all three for sanity checking.

\clearpage
\section{Additional Accuracy Results}
\label{app:accuracy}

\noindent\textbf{Accuracy on CatBoost and XGBoost.}
\Cref{tab:accuracy_all} reports the avg AUC and retention on CatBoost and XGBoost over the $16$ binary datasets. The results show that the strong retention observed for LightGBM extends to other tree-based students. CatBoost and XGBoost retain nearly all teacher AUC, with several settings exceeding $100\%$ retention. The best CatBoost student reaches $.877$ AUC, slightly above the best teacher in \Cref{tab:accuracy}, while the best XGBoost students reach $.875$ AUC. This supports the main-text conclusion that tree-based students are reliable distillation targets and that distillation can sometimes improve AUC rather than only compressing the teacher.
\begin{table}[h!]
\caption{Avg AUC and retention on CatBoost and XGBoost over the $16$ binary datasets.}
\label{tab:accuracy_all}
\centering
\resizebox{0.7\textwidth}{!}{%
\begin{tabular}{llcc}
\toprule
Type & Model & AUC & Ret. \\
\midrule
\multirow{5}{*}{\shortstack[l]{Distilled\\(CatBoost)}}
 & TabICLv2$\to$CB & \textbf{.876} $\pm$ .109 & 100.4\% \\
 & LimiX$\to$CB & .875 $\pm$ .111 & 100.0\% \\
 & TabPFNv2.6$\to$CB & .875 $\pm$ .112 & 100.5\% \\
 & TabDPT$\to$CB & .872 $\pm$ .113 & 100.5\% \\
 & Orion-MSP$\to$CB & .870 $\pm$ .110 & 100.7\% \\
\midrule
\multirow{4}{*}{\shortstack[l]{Distilled\\(CatBoost,\\multi-teacher)}}
 & {[}PFNv2.6+ICL+Limix+TabDPT{]}$\to$CB & \textbf{.877} $\pm$ .110 & -- \\
 & {[}PFNv2.6+ICL+Limix{]}$\to$CB & .876 $\pm$ .110 & -- \\
 & {[}PFNv2.6+Limix{]}$\to$CB & .876 $\pm$ .110 & -- \\
 & {[}PFNv2.6+OrionMSP+Limix{]}$\to$CB & .875 $\pm$ .110 & -- \\
\midrule
\multirow{5}{*}{\shortstack[l]{Distilled\\(XGBoost)}}
 & TabICLv2$\to$XGB & \textbf{.875} $\pm$ .109 & 100.3\% \\
 & TabPFNv2.6$\to$XGB & .874 $\pm$ .112 & 100.5\% \\
 & LimiX$\to$XGB & .874 $\pm$ .110 & 99.9\% \\
 & TabDPT$\to$XGB & .869 $\pm$ .113 & 100.2\% \\
 & Orion-MSP$\to$XGB & .868 $\pm$ .109 & 100.4\% \\
\midrule
\multirow{4}{*}{\shortstack[l]{Distilled\\(XGBoost,\\multi-teacher)}}
 & {[}PFNv2.6+ICL+Limix{]}$\to$XGB & \textbf{.875} $\pm$ .110 & -- \\
 & {[}PFNv2.6+ICL+Limix+TabDPT{]}$\to$XGB & .875 $\pm$ .110 & -- \\
 & {[}PFNv2.6+Limix{]}$\to$XGB & .875 $\pm$ .110 & -- \\
 & {[}PFNv2.6+OrionMSP+Limix{]}$\to$XGB & .874 $\pm$ .110 & -- \\
\bottomrule
\end{tabular}}
\end{table}

\noindent\textbf{Accuracy on multi-class datasets.} \Cref{tab:accuracy_multi_class} reports the avg AUC and retention of all students over the $3$ multi-class datasets. The results highlight that multiclass results are broadly consistent with the binary setting but with smaller differences between methods. Tree-based students retain high teacher performance, with LightGBM and XGBoost reaching up to $.786$--$.787$ macro-AUC and retention near or above $100\%$ for some teachers. MLP students remain slightly weaker, with retention between $96.9\%$ and $99.9\%$. These results suggest that distillation transfers reasonably well to multiclass tasks, although the small number of datasets makes the trend less conclusive than in the binary benchmark.

\begin{table}[h!]
\caption{Multiclass panel results (3 datasets, macro-AUC). Per-dataset mean, $\pm$ std across datasets.}
\label{tab:accuracy_multi_class}
\centering
\footnotesize
\begin{tabular}{llcc}
\toprule
Type & Model & AUC & Ret. \\
\midrule
\multirow{3}{*}{Baseline}
 & LogisticRegression & .785 $\pm$ .303 & -- \\
 & XGBoost & .770 $\pm$ .323 & -- \\
 & LightGBM & .770 $\pm$ .323 & -- \\
 \midrule
\multirow{6}{*}{Teacher}
 & TabDPT & \textbf{.789} $\pm$ .297 & -- \\
 & TabICLv2 & .788 $\pm$ .299 & -- \\
 & LimiX & .784 $\pm$ .304 & -- \\
 & TabPFNv2.6 & .782 $\pm$ .307 & -- \\
 & TabPFN (v2.5) & .779 $\pm$ .312 & -- \\
 & Orion-MSP v1.5 & .773 $\pm$ .319 & -- \\
\midrule
\multirow{5}{*}{\shortstack[l]{Distilled\\(CatBoost)}}
 & TabICLv2$\to$CB & \textbf{.782} $\pm$ .304 & 99.2\% \\
 & LimiX$\to$CB & .778 $\pm$ .312 & 99.2\% \\
 & Orion-MSP$\to$CB & .777 $\pm$ .313 & 100.5\% \\
 & TabDPT$\to$CB & .776 $\pm$ .314 & 98.4\% \\
 & TabPFNv2.6$\to$CB & .774 $\pm$ .320 & 98.9\% \\
\midrule
\multirow{5}{*}{\shortstack[l]{Distilled\\(XGBoost)}}
 & TabICLv2$\to$XGB & \textbf{.787} $\pm$ .299 & 99.9\% \\
 & TabDPT$\to$XGB & .780 $\pm$ .309 & 98.8\% \\
 & LimiX$\to$XGB & .777 $\pm$ .313 & 99.1\% \\
 & TabPFNv2.6$\to$XGB & .777 $\pm$ .315 & 99.3\% \\
 & Orion-MSP$\to$XGB & .776 $\pm$ .315 & 100.3\% \\
\midrule
\multirow{6}{*}{\shortstack[l]{Distilled\\(LightGBM)}}
 & TabICLv2$\to$LGBM & \textbf{.786} $\pm$ .299 & 99.8\% \\
 & TabDPT$\to$LGBM & .785 $\pm$ .302 & 99.5\% \\
 & Orion-MSP$\to$LGBM & .785 $\pm$ .301 & 101.5\% \\
 & LimiX$\to$LGBM & .784 $\pm$ .303 & 100.0\% \\
 & TabPFNv2.6$\to$LGBM & .782 $\pm$ .305 & 100.0\% \\
 & TabPFN (v2.5)$\to$LGBM & .779 $\pm$ .310 & 100.1\% \\
\midrule
\multirow{6}{*}{\shortstack[l]{Distilled\\(MLP)}}
 & TabPFN$\to$MLP & \textbf{.778} $\pm$ .311 & 99.9\% \\
 & LimiX$\to$MLP & .772 $\pm$ .319 & 98.5\% \\
 & TabPFNv2.6$\to$MLP & .772 $\pm$ .320 & 98.7\% \\
 & TabDPT$\to$MLP & .769 $\pm$ .323 & 97.5\% \\
 & Orion-MSP$\to$MLP & .766 $\pm$ .328 & 99.0\% \\
 & TabICLv2$\to$MLP & .763 $\pm$ .332 & 96.9\% \\
\bottomrule
\end{tabular}
\end{table}

\section{Additional Latency Results}
\label{app:latency}

\begin{figure}[h!]
\centering
\includegraphics[width=\textwidth]{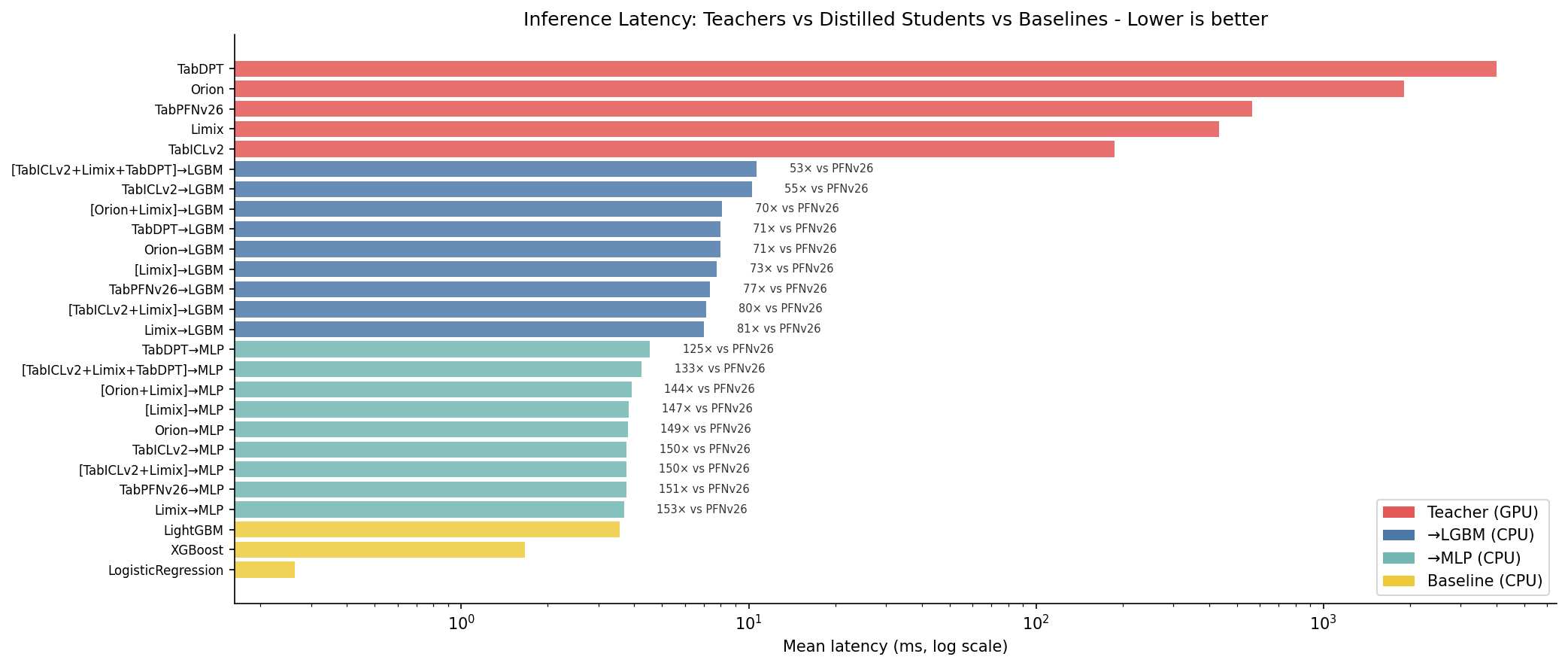}
\caption{Inference latency, log scale. Teachers (red) sit at 200--4000\,ms on GPU. Distilled students sit at 4--11\,ms on CPU; tree baselines below 4\,ms. Distillation collapses the latency gap.}
\label{fig:latency}
\end{figure}

Figure~\ref{fig:latency} and Table~\ref{tab:latency_all} confirm that the latency gains hold across all distilled settings. All MLP students run between $3.7$ and $4.5$\,ms, and all LGBM students between $7.0$ and $10.7$\,ms on CPU, while the fastest GPU-based teacher, TabICLv2, requires $187.4$\,ms. Even at P99, distilled students remain below $17$\,ms, whereas teachers range from $202.6$\,ms to over $4$\,s. The figure therefore highlights a consistent deployment gap: distillation moves inference from hundreds or thousands of GPU milliseconds to single-digit CPU milliseconds, with MLP offering the lowest latency and LGBM remaining within a practical real-time range.

\begin{table}[h!]
\caption{Inference latency and throughput, for all settings of distilled students, averaged across datasets ranging from 91 to 1821 test examples. All measurements on a single CPU core; teachers run on GPU.}
\label{tab:latency_all}
\centering
\resizebox{0.65\textwidth}{!}{%
\begin{tabular}{lrrr}
\toprule
Model & Mean ms & P99 ms & Throughput \\
\midrule
\multicolumn{4}{l}{\textit{Baselines}} \\
\midrule
LogisticRegression & \textbf{0.3} & 0.3 & 2{,}249K/s \\
XGBoost            & 1.7 & 2.7 & 353K/s \\
LightGBM           & 3.6 & 4.1 & 162K/s \\
\midrule
\multicolumn{4}{l}{\textit{TFM teachers (GPU)}} \\
\midrule
TabICLv2     & 187.4  & 202.6  & 3{,}221/s \\
LimiX        & 433.8  & 457.1  & 1{,}450/s \\
TabPFNv2.6   & 564.9  & 577.8  & 1{,}099/s \\
Orion-MSP v1.5 & 1{,}911.9 & 1{,}941.1 & 323/s \\
TabDPT       & 4{,}010.0 & 4{,}024.1 & 600/s \\
\midrule
\multicolumn{4}{l}{\textit{Distilled MLP students}} \\
\midrule
LimiX$\to$MLP                                & 3.7 & 4.0 & 172K/s \\
TabPFNv2.6$\to$MLP                           & 3.8 & 4.1 & 170K/s \\
TabICLv2$\to$MLP                             & 3.8 & 4.4 & 169K/s \\
{[}PFNv2.6+Limix{]}$\to$MLP                  & 3.8 & 4.8 & 166K/s \\
{[}PFNv2.6+ICL+Limix{]}$\to$MLP              & 3.8 & 4.3 & 166K/s \\
Orion-MSP$\to$MLP                            & 3.8 & 4.5 & 164K/s \\
{[}PFNv2.6+OrionMSP+Limix{]}$\to$MLP         & 3.9 & 4.7 & 157K/s \\
{[}PFNv2.6+ICL+Limix+TabDPT{]}$\to$MLP       & 4.2 & 4.7 & 150K/s \\
TabDPT$\to$MLP                               & 4.5 & 5.2 & 158K/s \\
\midrule
\multicolumn{4}{l}{\textit{Distilled LGBM students}} \\
\midrule
LimiX$\to$LGBM                               & 7.0 & 14.2 & 87K/s \\
{[}PFNv2.6+ICL+Limix{]}$\to$LGBM             & 7.1 & 12.5 & 92K/s \\
TabPFNv2.6$\to$LGBM                          & 7.3 & 12.5 & 82K/s \\
{[}PFNv2.6+Limix{]}$\to$LGBM                 & 7.7 & 13.1 & 98K/s \\
TabDPT$\to$LGBM                              & 8.0 & 12.6 & 79K/s \\
Orion-MSP$\to$LGBM                           & 8.0 & 10.0 & 86K/s \\
{[}PFNv2.6+OrionMSP+Limix{]}$\to$LGBM        & 8.1 & 15.7 & 78K/s \\
TabICLv2$\to$LGBM                            & 10.3 & 14.6 & 67K/s \\
{[}PFNv2.6+ICL+Limix+TabDPT{]}$\to$LGBM      & 10.7 & 16.1 & 64K/s \\
\bottomrule
\end{tabular}}
\end{table}

\section{Multi-teacher analysis with full coverage}
\label{app:multi}

\Cref{tab:multi} reports multi-teacher LGBM distillation on the full 19-dataset benchmark. None of the four ensembles that ran on all datasets exceeds the best single-teacher student (TabICLv2$\to$LGBM at 0.906); they are tied at three decimal places. On the same subset, the best single teacher scores 0.916, so there is some real ensemble effect on those three. Whether it generalises requires running the ensemble on the remaining ten datasets, which we will do in a follow-up.

\begin{table}[h!]
\caption{Multi-teacher LGBM distillation across binary datasets. The first four rows ran on all 16 datasets; the last ran on 3.}
\label{tab:multi}
\centering
\footnotesize
\begin{tabular}{lcc}
\toprule
Teacher ensemble & AUC (mean) & Coverage (datasets) \\
\midrule
$[$TabPFN $+$ Limix$]$ & .906 & 16 \\
$[$TabPFN $+$ TabICLv2 $+$ Limix$]$ & .906 & 16 \\
$[$TabPFN $+$ OrionMSP $+$ Limix$]$ & .906 & 16 \\
$[$TabPFN $+$ TabICLv2 $+$ Limix $+$ TabDPT$]$ & .906 & 16 \\
\midrule
Best single-teacher LGBM (TabICLv2$\to$LGBM) & .906 & 16 \\
Best single teacher (TabICLv2) & .918 & 16 \\
\bottomrule
\end{tabular}
\end{table}

\end{document}